\documentclass[10pt,twocolumn,letterpaper]{article}

\usepackage[pagenumbers]{cvpr} 

\usepackage[dvipsnames]{xcolor}

\usepackage{graphicx,scalerel}
\usepackage{amsmath}
\usepackage{amssymb}
\usepackage{booktabs}

\newcommand{\hide}[1]{}

\usepackage{color}
\usepackage{xcolor,colortbl}
\usepackage{mathtools}
\usepackage{multirow}
\usepackage{bm} 
\definecolor{purple}{rgb}{0.65,0,0.65}
\definecolor{dark_green}{rgb}{0, 0.5, 0}
\definecolor{blueish}{rgb}{0.0, 0.3, .6}
\definecolor{tabhighlight}{HTML}{e5e5e5}
\newcolumntype{h}{>{\columncolor{tabhighlight}}c}

\makeatletter
\newcommand*\wt[2][0.15ex]{%
        \begingroup
        \mathchoice{\wt@helper{#1}{#2}{\displaystyle}{\textfont}}
                   {\wt@helper{#1}{#2}{\textstyle}{\textfont}}
                   {\wt@helper{#1}{#2}{\scriptstyle}{\scriptfont}}
                   {\wt@helper{#1}{#2}{\scriptscriptstyle}{\scriptscriptfont}}%
        \endgroup
        #2%
}
\newcommand*\wt@helper[4]{%
        \def\currentfont{\the#41}%
        \def\currentskewchar{\char\the\skewchar\currentfont}%
        \setbox\tw@\hbox{\currentfont#2\currentskewchar}%
        \dimen@ii\wd\tw@
        \setbox\tw@\hbox{\currentfont#2{}\currentskewchar}%
        \advance\dimen@ii-\wd\tw@
        \rlap{\raisebox{-#1}{$\m@th#3\kern\dimen@ii\widetilde{\phantom{#2}}$}}%
}
\makeatother

\usepackage[export]{adjustbox} 

\newlength{\gridimagewidth}
\usepackage{array}
\newcolumntype{C}[1]{>{\centering\let\newline\\\arraybackslash\hspace{0pt}}m{#1}}

\newcolumntype{h}{>{\columncolor{tabhighlight}}c}

\newlength{\syntheticwidth}
\newlength{\realworldwidth}
\definecolor{cvprblue}{rgb}{0.21,0.49,0.74}
\usepackage[pagebackref,breaklinks,colorlinks]{hyperref}

\usepackage[capitalize]{cleveref}
\crefname{section}{Sec.}{Secs.}
\Crefname{section}{Section}{Sections}
\Crefname{table}{Table}{Tables}
\crefname{table}{Tab.}{Tabs.}

\def\paperID{*****} 
\def\confName{CVPR}
\def\confYear{2024}

\title{Restoration of Analog Videos Using Swin-UNet}

\author{Lorenzo Agnolucci \and Leonardo Galteri \and  Marco Bertini \and Alberto Del Bimbo \vspace{0.1ex} \and
University of Florence - Media Integration and Communication Center (MICC) \\
Florence, Italy\\
{\tt\small [name.surname]@unifi.it}
}

\begin{document}

\begin{figure}[htb]
\twocolumn[{
\renewcommand\twocolumn[1][]{#1}%
\maketitle
\vspace{-25pt}
\begin{center}
\begin{subfigure}{0.3265\linewidth}
        \includegraphics[width=\linewidth]{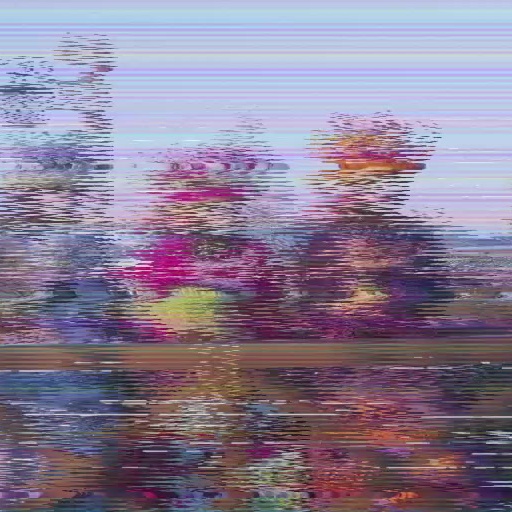}
        \caption{Input}
    \end{subfigure}
    \hfill
    \begin{subfigure}{0.3265\linewidth}
        \includegraphics[width=\linewidth]{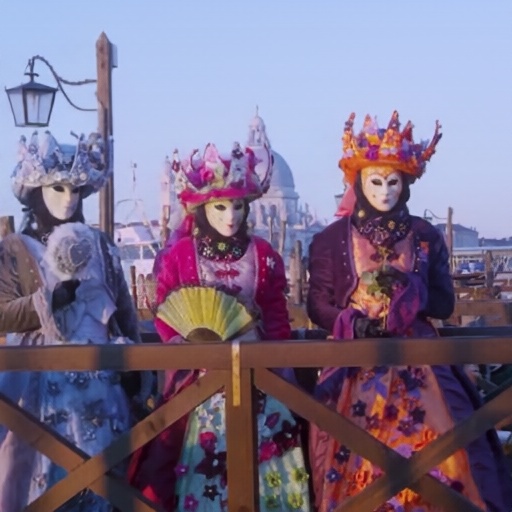}
        \caption{Restored}
    \end{subfigure}
    \hfill
    \begin{subfigure}{0.3265\linewidth}
        \includegraphics[width=\linewidth]{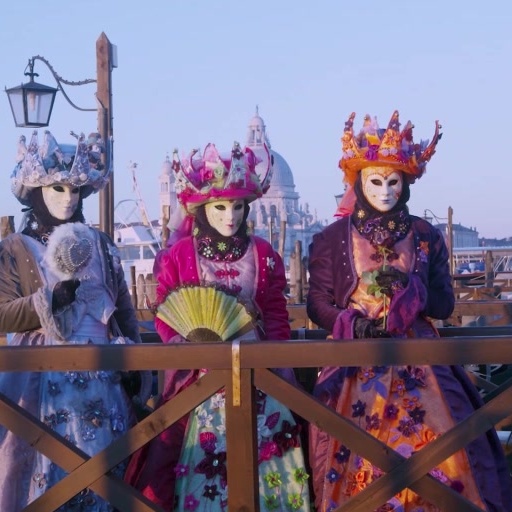}
        \caption{Ground Truth}
    \end{subfigure}
\end{center}
\vspace{-0.5cm}
\caption{Qualitative results of video restoration on the synthetic dataset}
\label{fig:teaser}
\vspace{0.5cm}
}]
\end{figure}

\begin{abstract}
In this paper, we present a system to restore analog videos of historical archives. These videos often contain severe visual degradation due to the deterioration of their tape supports that require costly and slow manual interventions to recover the original content. The proposed system uses a multi-frame approach and is able to deal with severe tape mistracking, which results in completely scrambled frames. Tests on real-world videos from a major historical video archive show the effectiveness of our demo system. The code and the pre-trained model are publicly available at \small{\href{https://github.com/miccunifi/analog-video-restoration}{\url{https://github.com/miccunifi/analog-video-restoration}}}.
\end{abstract}    
\section{Introduction and Related Work}
\label{sec:intro}

Historical videos constitute an important part of the cultural heritage of a society. This content often is hampered by numerous artifacts and degradations due to technological limitations and aging of the recording support that limit its distribution and fruition by the general public. Normally the restoration of these videos is conducted frame by frame by experienced archivists with commercial solutions, thus at great economic and time cost.

For this reason, some works tried to restore historical video archives more rapidly and without human aid. \cite{deoldify} is an open-source tool for old films restoration that presents a NoGAN training approach. \cite{iizuka2019deepremaster} relies fully on 3D convolutions and on source-reference attention for frame colorization. \cite{wan2022oldfilm} proposes a recurrent transformer network that localizes defects in an unsupervised manner. 

\textit{Istituto Luce Cinecittà}, an Italian society responsible for the preservation and distribution of the \textit{Archivio Storico Luce}, the largest Italian historical video archive dating from throughout the 1900s and comprising a variety of sources, provided us with some analog videos from this archive. These videos present several system-intrinsic and aging-related types of degradations typical of analog video tapes. The related works focus on standard structured defects such as scratches and cracks, so they are not capable of restoring the particular types of artifacts that these videos present. Unfortunately, being real-world videos, there is no clean high-quality version of them to use as ground truth for supervised learning. Consequently, we created a synthetic dataset as similar as possible to the real-world videos to train our system.

\section{Proposed Approach}

\subsection{Synthetic Dataset} \label{sec:synthetic_dataset}

In order to train a restoration model, we created a synthetic dataset as similar as possible to the real-world videos. Starting from high-quality videos of the Harmonic dataset \cite{Harmonic-2019} we used Adobe After Effects \cite{christiansen2013adobe} to randomly add several types of degradations, such as:
\begin{itemize}
    \item Gaussian noise, resembling the tape noise that is typical of analog videos;
    \item white artifacts simulating tape dropouts;
    \item cyan, magenta and green horizontal lines resembling chroma fringing;
    \item horizontal displacements, similar to tape mistracking artifacts; this is the most complex error that can be encountered.
\end{itemize}
As with the real-world videos, all these artifacts vary over time and occur with different intensities, positions, and combinations for each frame.

We ended up with 26392 frames that we divided into training and validation sets with an 80-20 ratio. Then this synthetic dataset was used to train the model described in section  \ref{sec:network_architecture}.

\subsection{Network Architecture} \label{sec:network_architecture}
\begin{figure}
    \captionsetup{skip=5pt}
    \centering
    \includegraphics[width=\linewidth]{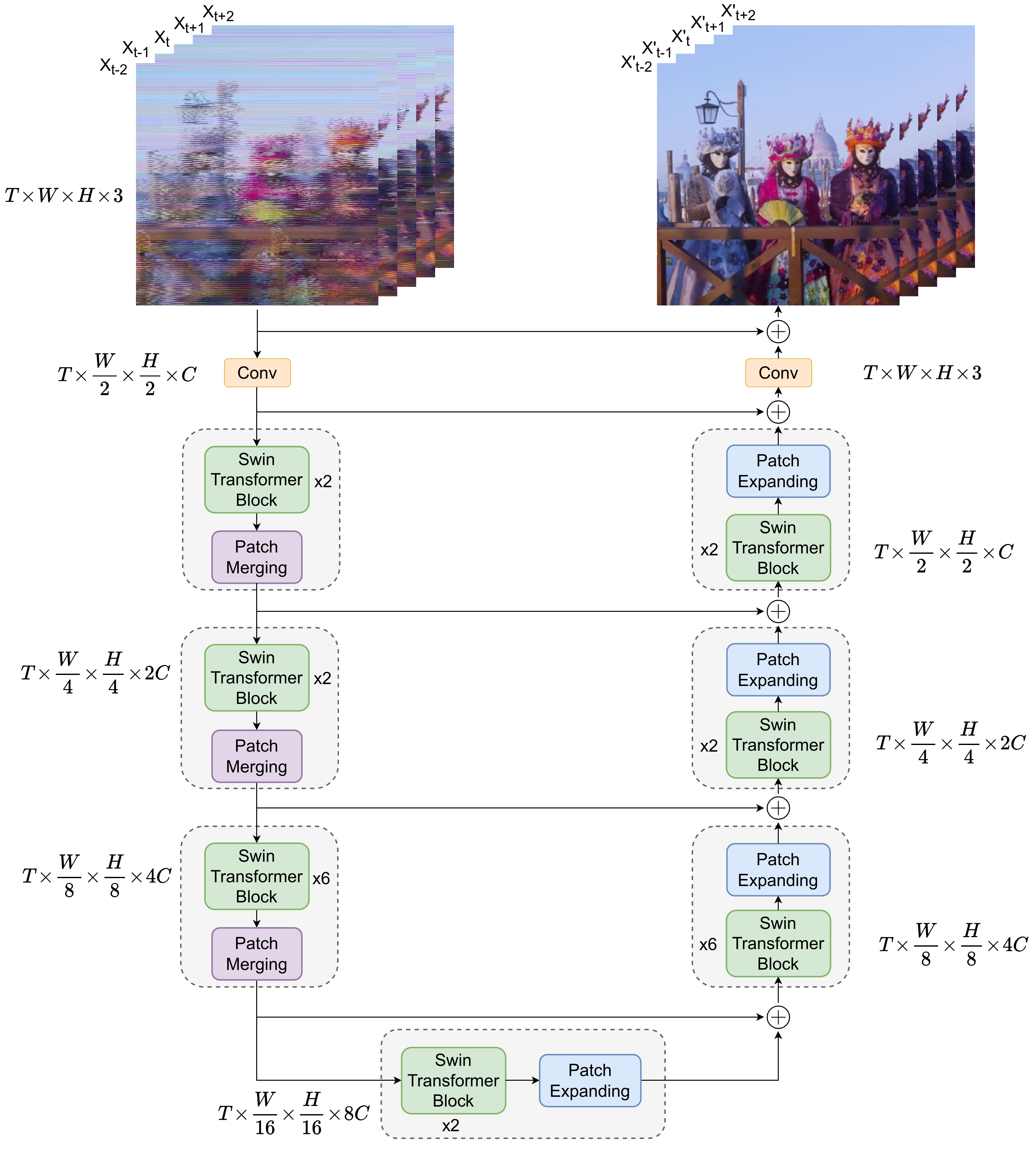}
    \caption{Proposed network architecture.}
    \label{fig:architecture}
    \vspace{0.3cm}
\end{figure}

Inspired by \cite{cao2021swin}, we developed a Swin-UNet architecture presented in Figure \ref{fig:architecture}. Differently from \cite{cao2021swin} our network works on videos, so we converted it to a multi-frame approach. In this way, the model enhances T frames at once exploiting spatio-temporal information. Moreover, we employed 3D convolution for partitioning the input into patches and pixel shuffle for the patch expanding layer. Following \cite{galteri2020increasing}, a skip connection between the degraded input and restored output makes the network learn the residual of each frame. This choice reduces the overall training time and improves its stability.

The training loss is a weighted sum of a pixel loss (in particular, the Mean Square Error) and a perceptual loss \cite{johnson2016perceptual, ledig2017photo, dosovitskiy2016generating} defined on the VGG-19 \cite{simonyan2014very} feature space. The network was trained with $256\times256$ patches cropped randomly from the input frames. During training and testing the number of frames T processed by the model was fixed to 5.

\section{Experimental Results}

\begin{table}
    \centering
    \Huge
    \resizebox{0.75\linewidth}{!}{ 
    \begin{tabular}{lccc}
        \toprule
        Method & PSNR$\uparrow$ & SSIM$\uparrow$ & LPIPS$\downarrow$ \\
        \midrule
        DeOldify \cite{deoldify} & 11.56 & 0.451 & 0.671 \\
        \textbf{Ours} & \textbf{34.78} & \textbf{0.939} & \textbf{0.063} \\
        \bottomrule
    \end{tabular}}
    \caption{Quantitative results on the synthetic dataset}
    \label{tab:quantitative_results}
\end{table}

We measured the performance of our method using three standard full-reference visual quality metrics: 1) PSNR; 2) SSIM \cite{wang2004image}; 3) LPIPS \cite{zhang2018unreasonable}. The quantitative results obtained for the restoration of the $512\times512$ central crop on the synthetic dataset are reported in Tab.~\ref{tab:quantitative_results}.
For a fair comparison, we re-trained DeOldify \cite{deoldify} from scratch using our training data. Our model achieves the best performance.

\begin{figure}
    \captionsetup{skip=5pt}
    \centering
    \begin{minipage}[b]{.48\linewidth}
    \includegraphics[width=\textwidth]{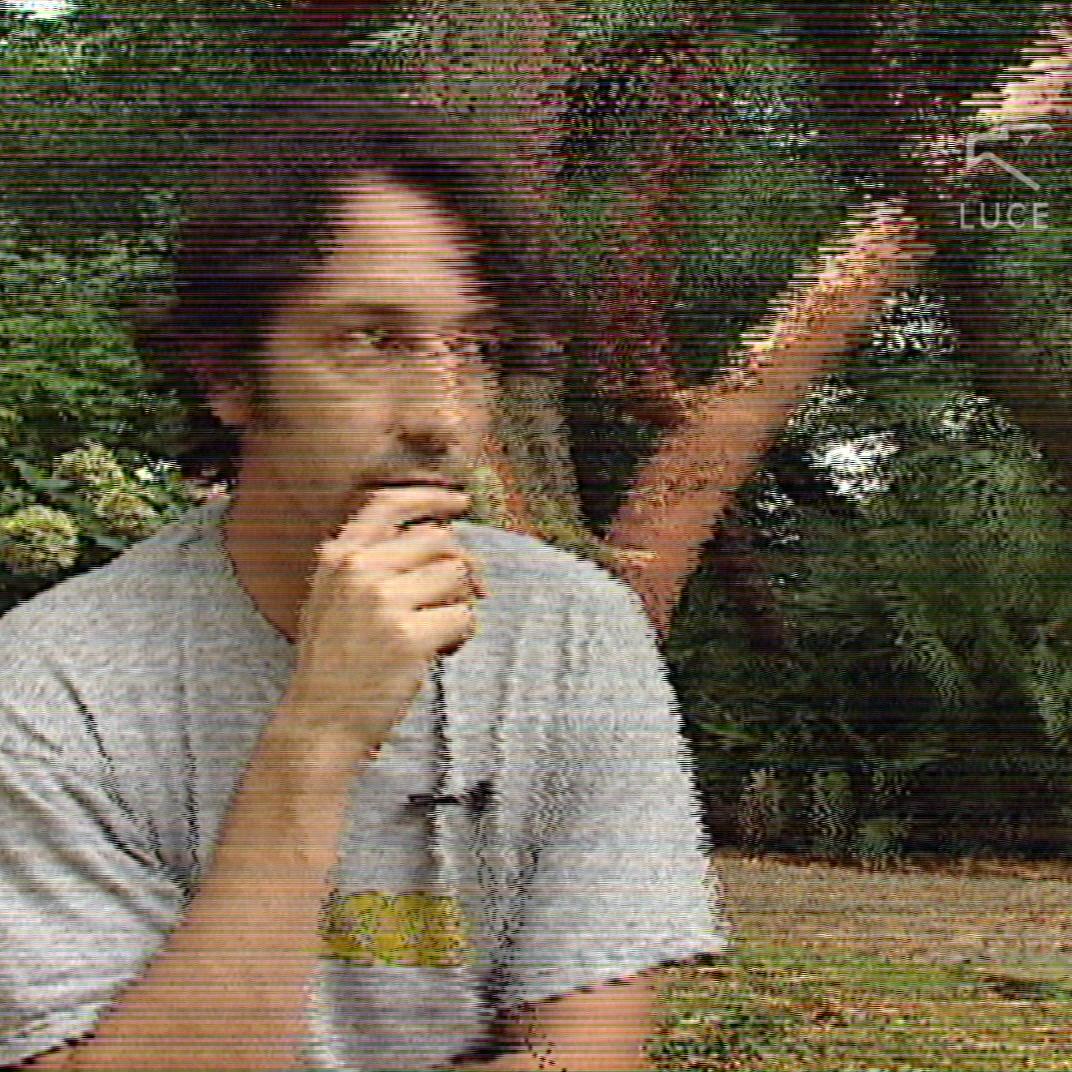}
    \caption*{(a): Input}
    \end{minipage}
    \hfill
    \begin{minipage}[b]{.48\linewidth}
    \includegraphics[width=\textwidth]{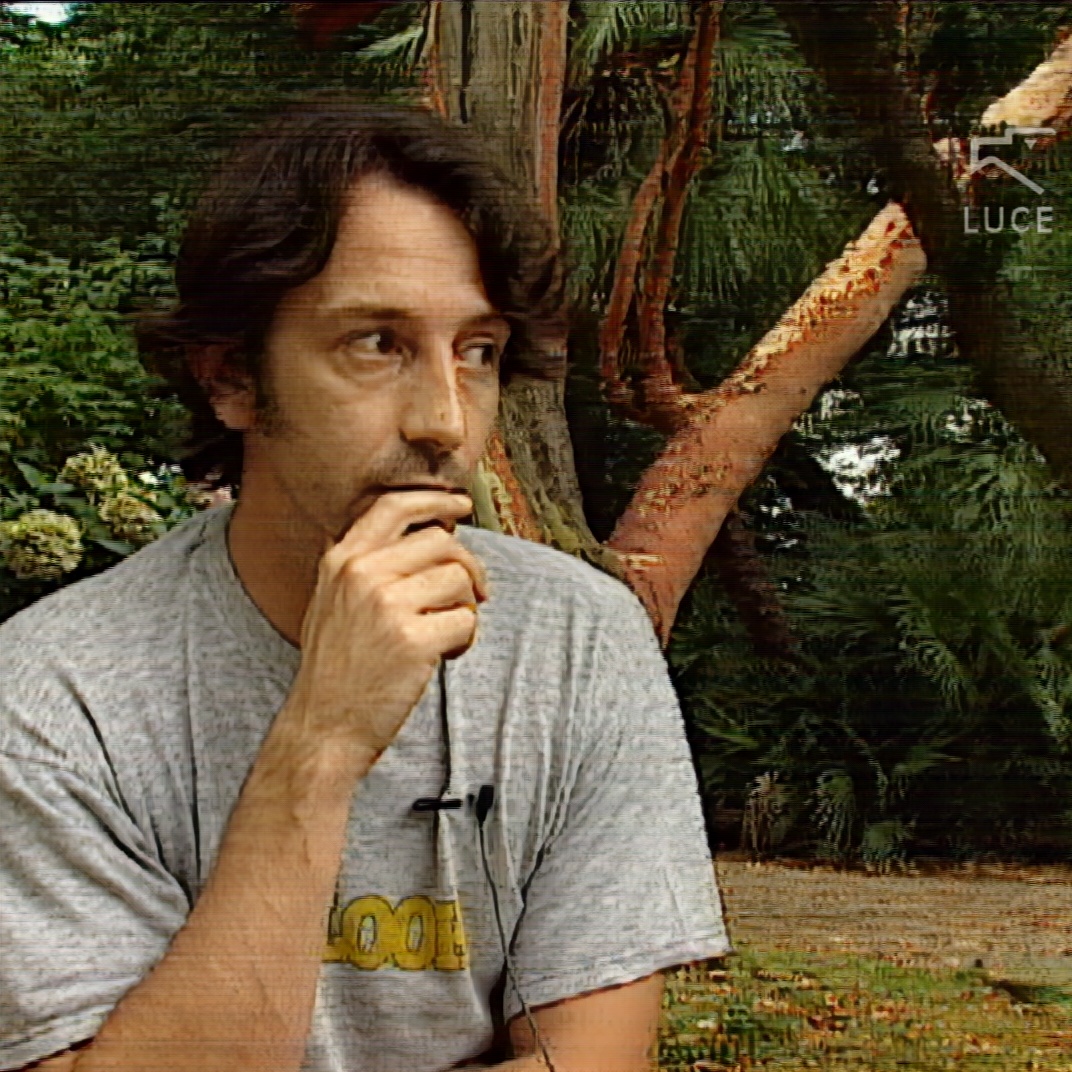}
    \caption*{(b): Restored}
    \end{minipage}
    \caption{Qualitative results on the real-world video of the test dataset}
    \label{fig:results_original}
\end{figure}

The qualitative results for the synthetic and real-world datasets are presented in Figure \ref{fig:teaser} and \ref{fig:results_original}, respectively. Our model proves to be able to restore a lot of details lost with the heavy degradations to which the input frames had been subjected. Indeed, thanks to the spatio-temporal information captured by our multi-frame approach the network can exploit the time-varying nature of the artifacts and address the most severe tape mistracking.

To let users restore degraded videos with similar artifacts we developed a Flask-based demo web app accessible through a web browser. Our platform supports the upload of video files and provides the user with the downloadable restored result, as well as a comparison with the original video. Alternatively, the user can choose one of our example videos just to see what our model is capable of.

\section{Conclusion}
In this work, we focused on restoring analog videos of historical archives. We created a synthetic dataset to train the Swin-UNet network we designed. Tests on synthetic and real-world videos prove the effectiveness of our approach. We also developed a demo web app to let users restore videos with similar artifacts.

\paragraph{Acknowledgments}
This work was partially supported by the European Commission under European Horizon 2020 Programme, grant number 101004545 - ReInHerit.
{
    \small
    \bibliographystyle{ieeenat_fullname}
    \bibliography{main}
}


\end{document}